%% file: naaclhlt2018.tex
\documentclass[11pt,a4paper,dvipsnames,hidelinks]{article}
\usepackage[hyperref]{naaclhlt2018}
\usepackage{times}
\usepackage{latexsym}
\usepackage{graphicx}
\usepackage{array}
\usepackage{url}
\usepackage{cleveref}
\usepackage{blindtext}
\usepackage{xspace}
\usepackage{subfiles}
\usepackage{multirow}
\usepackage{booktabs}
\usepackage{bm}
\usepackage{enumitem}

\newcommand{\flagged}{\textit{flagged}\xspace}
\newcommand{\green}{\textit{green}\xspace}
\aclfinalcopy 
\def\aclpaperid{115} 

\newcommand{\lcb}{{\tt {\char '173}}}   
\newcommand{\rcb}{{\tt {\char '175}}}   

\title{
  Helping or Hurting? Predicting Changes in Users' Risk of \\
  Self-Harm Through Online Community Interactions
}

\author{Luca Soldaini*, Timothy Walsh$\bm{^\dagger}$, Arman Cohan*, Julien Han$\bm{^\dagger}$, Nazli Goharian*\\
  {*}Information Retrieval Lab, Georgetown University\\
  $^\dagger$Georgetown University \\
  {*}{\tt \lcb luca, arman, nazli\rcb @ir.cs.georgetown.edu}\\
  $^\dagger${\tt \lcb tw614, mh1795\rcb @georgetown.edu}}

\date{}

\begin{document}
\maketitle
\begin{abstract}
In recent years, online communities have formed around suicide and self-harm prevention.
While these communities offer support in moment of crisis, they can also normalize harmful behavior, discourage professional treatment, and instigate suicidal ideation.
In this work, we focus on how interaction with others in such a community affects the mental state of users who are seeking support.
We first build a dataset of conversation threads between users in a distressed state and community members offering support. We then show how to construct a classifier to predict whether distressed users are helped or harmed by the interactions in the thread, and we achieve a macro-F1 score of up to 0.69.
\end{abstract}

\subfile{sections/introduction}
\subfile{sections/related}
\subfile{sections/dataset}
\subfile{sections/methods}
\subfile{sections/results}
\subfile{sections/conclusions}

\appendix

\end{document}

%% file: sections/introduction.tex
\documentclass[../naaclhlt2018.tex]{subfiles}

\section{Introduction}

Suicide is a major challenge for public health.
Worldwide, suicide is the 17th leading cause of death, claiming 800,000 lives each year \cite{world2015world}.
In the United States alone, 43,000 Americans died from suicide in 2016 \cite{afsp2016suicide}, a 30-year high \cite{tavernise2016us}.

In recent years, online communities have formed around suicide and self-harm prevention.
The Internet offers users an opportunity to discuss their mental health and receive support more readily while preserving privacy and anonymity \cite{robinson2016social}.
However, researchers have also raised concerns about the effectiveness and safety of online treatments \cite{robinson2015safety}.
In particular, a major concern is that, through interactions with each other, at-risk users might normalize harmful behavior \cite{daine2013power}.
This phenomenon, commonly referred to as the ``Werther effect,'' has been amply studied in psychology literature and observed across various cultures and time periods \cite{phillips,gladwell,cheng,niederkrotenthaler2009copycat}.
In the Natural Language Processing (NLP) community, computational methods have been used to study how high profile suicides influence social media users \cite{dechoudhury2013predicting,kumar2015detecting} and study the role of empathy in counseling \cite{perez2017understanding} and online health communities \cite{khanpour2017identifying,de2017language}.
However, most studies about contagious suicidal behavior in online communities are small-scale and qualitative \cite{haw2012suicide,hawton2014self}.

\begin{figure}[t]
    \centering
    \includegraphics[width=0.8\columnwidth]{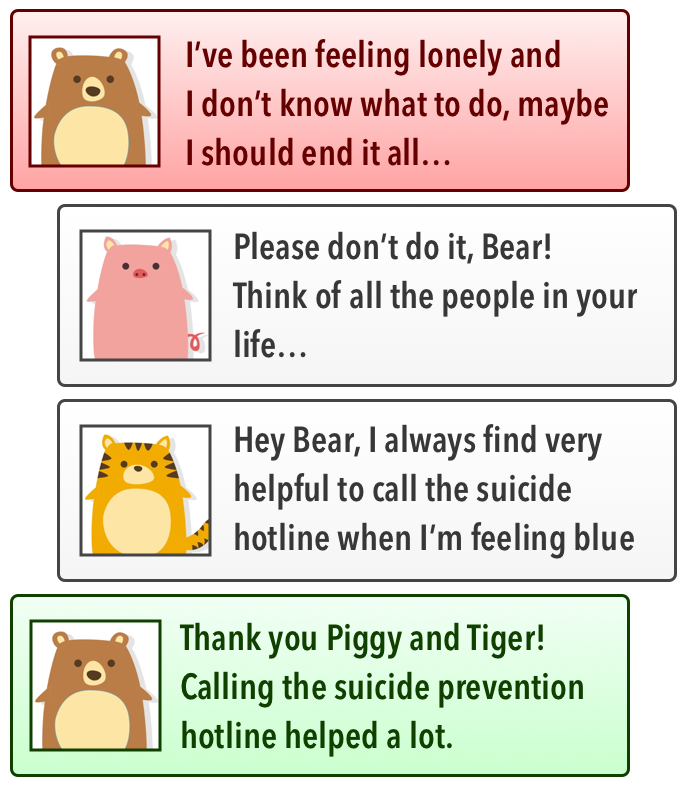}
    \caption{A fictitious example of \flagged thread with final \textit{green} label (we avoid publishing any text from the dataset in order to preserve users' privacy.)}
    \label{distr_threads}
\end{figure}

In this work, we set out to study how users affect each other through interactions in an online mental health support community.
In particular, we focus on users who are in a distressed or suicidal state and open a conversation thread to seek support.
Our goal is to model how such users respond to interactions with other members in the community.
First, we extract conversation threads by combining the initial post from the distressed user and the set of replies from other users who participate in the discussion.
All conversation threads in our dataset are from the 2016 Computational Linguistics and Clinical Psychology (CLPsych) Workshop's ReachOut.com dataset \cite{milne}, a collection of 64,079 posts from a popular support forum with more than 1.8 million annual
visits.
Then, we build a classifier that, given a thread, predicts whether the user at risk of self-harm or suicide successfully overcomes their state of distress through conversations with other community members.
The proposed system achieves up to a 0.69 macro-F1 score.
Furthermore, we extract and analyze significant features to explain what language may potentially help users in distress and what topics potentially cause harm.
We observe that mentions of family, relationships, religion, and counseling from community members are associated with a reduction of risk, while target users who express distress over family or work are less likely to overcome their state of distress.
Forum moderators and clinicians could deploy a system based on this research to highlight users experiencing a crisis, and findings could help train community members to respond more effectively.

In summary, our contribution is three-fold:
\begin{itemize}[wide, labelindent=0pt, labelwidth=!]
    \item We introduce a method for extracting conversation threads initiated by users in psychological distress from the 2016 CLPsych ReachOut.com dataset;
    \item We construct a classifier to predict whether an at-risk user can successfully overcome their state of distress through conversations with other community members;
    \item We analyze the most significant features from our model to study the language and topics in conversations with at-risk users.
\end{itemize}

%% file: sections/related.tex
\documentclass[../naaclhlt2018.tex]{subfiles}

\section{Related Work}

\subsection{Social media and suicide}
There is a close connection between natural language and mental-health, as language is an essential lens through which mental-health conditions can be examined \cite{coppersmith2017scalable}.
At the same time, due to the ubiquity of social media in the recent decades, a huge amount of data has become available for researchers to look at mental health challenges more closely.
Suicide and self-harm, which are among the most significant mental health challenges, have been recently studied through analyzing language in social media \cite{jashinsky2014tracking,thompson2014,gunn2015twitter,de2016discovering,coppersmith2016exploratory,conway2016social}
These works exploit various NLP methods to study and quantify the mental-health language in social media.
For example, \citet{coppersmith2015quantifying} focused on quantifying suicidal ideation among Twitter users. Through their experiments, they demonstrated that neurotypical and suicidal users can be separated when controlling for age and gender based on the language used on social media.
\citet{huang2015topic} combined topic modeling, distributional semantics, and specialized lexicon to identify suicidal users on social media.
Recently, CLPsych \cite{CLPsych:2016,CLPsych:2017} introduced shared tasks to identify the risk of suicide and self-harm in online forums. Through these shared tasks, participants explored various NLP methods for identifying users that are at risk of suicide or self-harm \cite{milne}. Most of the related work in this area uses variations of linear classifiers with features that quantify the language of users in social media. For example, \citet{kim-EtAl:2016:CLPsych} used a combination of bag-of-words and doc2vec feature
representations of the target forum posts. Their system achieved the top score, a macro-average F1 score of 0.42 over four levels of risk. Another successful system utilized a stack of feature-rich random forest and linear support vector machines \cite{malmasi-zampieri-dras:2016:CLPsych}. Finally, \citet{cohan2015triaging} used various additional contextual and psycholinguistic features.
In a follow-up work, \citet{cohan2017triaging} further improved the results on this dataset by introducing additional features and an ensemble of classifiers.
In addition to these methods, automatic feature extraction methods have also been explored to quantify suicide and self-harm \cite{yates2017}.
In this work, instead of directly assessing a user's risk level based on their own posts, we study how the language of other users affects the level of risk.

\subsection{Peer interaction effect on suicide}

Beside messages and individuals, researchers have long been interested in how individuals prone to suicidal ideation affect each other.
The most prominent examples in this area focused on examining the so-called ``Werther effect,'' i.e. the hypothesis that suicides or attempts that receive press coverage, or are otherwise well-known, cause copycat suicides.
Some of the earliest work related to our line of inquiry comes from the sociologist David Phillips who in the 1970s identified an increase in the suicide rate of American and British populations following newspaper publicity of a suicide, and argued that the relationship between publicity and the increase was causal \cite{phillips}. Other researchers later found similar results in other parts of the world \cite{gladwell,cheng}, across various types of media, and involving different types of subjects such as fictional characters and celebrities \cite{stack,niederkrotenthaler2009copycat}.

More recently, researchers have focused on studying instances of the Werther effect in online communities.
\citet{kumar2015detecting} examined posting activity and content after ten high-profile suicides and noted considerable changes indicating increased suicidal ideation.
In another work, \citet{de2016discovering} performed a study on Reddit users to infer which individuals might undergo transitions from mental health discourse to suicidal ideation.
They observe that suicidal ideation is associated with some psychological states such as heightened self-attentional focus and poor linguistic coherence.
In a subsequent work, they analyzed the influence of comments in Reddit mental health communities on suicidal ideation to establish a framework for identifying effective social support \citep{de2017language}.
In contrast with this work, we focus on studying peer influence on suicidal and self-harm ideation.
Online mental health-related support forums, being inherently discussion-centric, are an appropriate platform to investigate peer influence on suicidal ideation.

While most works on the Werther effect focus on passive exposure to print, broadcast, or online media across large populations, other research has studied the contagion of suicide within smaller social clusters \cite{hawton2014self,haw2012suicide}.
Recently, researchers have observed similar behavior in online communities \cite{daine2013power}.
However, research in this area tends to be qualitative rather than quantitative, thus ignoring the possibility of leveraging linguistic signals to prevent copycat suicides.

Finally, computational linguists have also investigated the use of empathy in counseling \cite{perez2017understanding} and online health communities \cite{khanpour2017identifying}. Both works focused on classification of empathetic language.
In the first, linguistic and acoustic features are used to identify high and low empathy in conversations between therapists and clients.
The second leverages two neural models (one convolutional, the other recurrent) to identify empathetic messages in online health communities; the two models are combined to achieve a 0.78 F1 score in detecting empathetic messages.
Unlike this work, their model focuses on predicting how empathy affects the next message from an at-risk user rather than modeling the entire conversation.

%% file: sections/dataset.tex
\documentclass[../naaclhlt2018.tex]{subfiles}

\section{Identifying Flagged Threads}
\label{sec:dataset}

\subsection{Methodology} 
\label{sub:methodology}



To study the effect of peer interaction on mental health and suicidal ideation, we leverage conversation threads from the 2016 CLPsych ReachOut.com dataset \citep{milne}. ReachOut.com is a popular and large mental health support forum with more than 1.8 million annual visits \cite{millen2015annual} where users engage in discussions and share their thoughts and experiences. In online forums, users typically start a conversation by creating a discussion thread. Other community members, including moderators, may then reply to the initial post and other replies in the thread. The post contents can be categorized into two severity categories, \flagged and \green \cite{milne}. The \flagged category means that the user might be at risk of self-harm or suicide and needs attention, whereas the \green category specifies content that does not need attention. Our goal is to investigate how the content of users changes over the course of discussion threads they initiate.

In particular, we focus on threads in which the first post was marked as \flagged, and use subsequent replies to predict the change in status for the user who initiated the thread.
More formally, let $t_i$ be a thread of posts $\{p_{i_0}, \ldots, p_{i_m}\}$, $u_T$ the user who initiated the thread $t_i$ (which we will refer to as ``target user'' for the reminder of this paper), and $\{u_{P_1}, \ldots, u_{P_o}\}$ users who reply in thread $t_i$ (we will refer to these users as ``participating users'').
Let $l_{i_j}$ be the label for post $p_{i_j}$, where $l_{i_j} = \textit{flagged}$ if the author of the post is in a distressed mental state and requires attention, and $l_{i_j} = \textit{green}$ if the author is not in a state of distress.
Given a post $p_{i_j}$ in $t_i$ authored by target user $u_T$ who is initially in a distressed state (indicated by  $l_{i_0} = \textit{flagged}$), our goal is to predict the state of $u_T$ at post $p_{i_j}$ (i.e., we want to predict $l_{i_j}$).

\begin{figure}[t]
    \centering
    \includegraphics[width=.98\columnwidth]{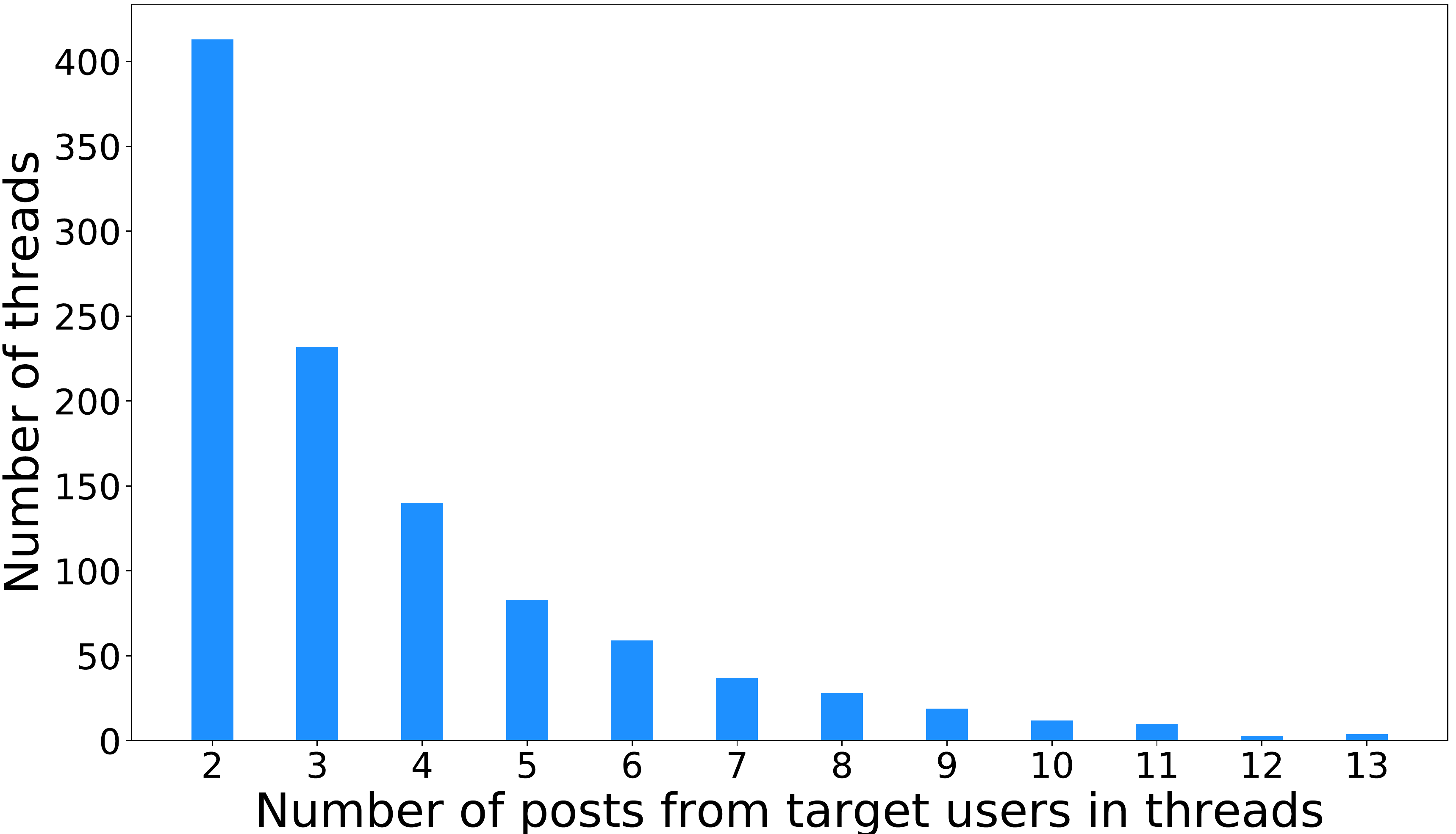}
    \caption{Distribution of threads in the dataset with respect to the number of times the target user posted in their own thread. The majority of threads contain two or three posts from the target user (the initial flagged posts and up to two replies.)}
    \label{counts_threads}
\end{figure}

Because of this experimental setting and the limited manual annotation of the CLPsych 2016 dataset, we cannot exclusively leverage the annotations provided with the dataset.
In fact, only 42 conversation threads with an initial post marked as \flagged can be extracted using the 1,227 manually-annotated posts.
Therefore, we supplement the manual annotations with automatic labels extracted using the high performance system described by \citet{cohan2017triaging}\footnote{The annotations for all posts in ReachOut.com were graciously provided by the authors of this work.}.
This system achieves an F1 score of 0.93 identifying \flagged vs \green posts on the CLPsych 2016 test set. Such high performance makes this system appropriate for annotating all posts as either \textit{flagged} or \textit{green} without the need for additional manual annotation.
To do so, we first obtain the probability of \textit{green} being assigned by the system to each of the 64,079 posts in the dataset, and then label as \textit{green} all posts whose probability is greater than threshold $\tau = 0.7751$.
We choose this value of $\tau$ because it achieves 100\% recall on \flagged posts in the 2016 CLPsych test set (therefore minimizing the number of users in emotional distress who are classified as \green), while still achieving high precision ($0.91$, less than 1\% worse than the result reported by \citet
{cohan2017triaging}.)

\begin{figure}[t]
    \centering
    \includegraphics[width=.98\columnwidth]{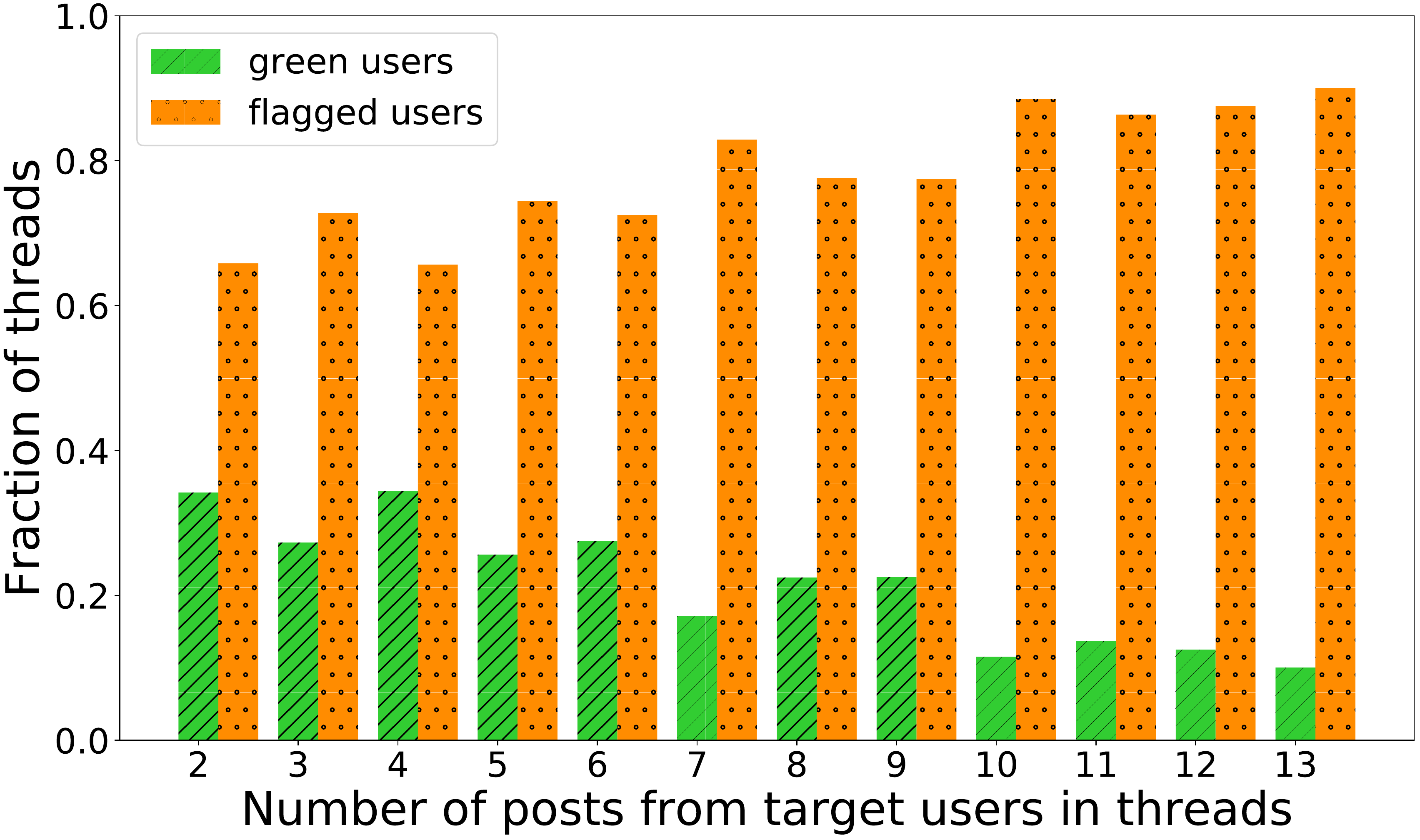}
    \caption{Distribution of labels from the final status of target users with respect to the number of times the target user posted in the thread. We observe a strong negative and statistically significant correlation between the number of target user posts and the likelihood of a final status of \textit{green} (Pearson's correlation, $\rho=-0.91$, $p<0.0001$.)}
    \label{distr_threads}
\end{figure}

Using the heuristic described above with the automatic post labels, we obtain 1,040 threads, each containing between 2 and 13 posts from the target user, including the initial post (mean=3.67, median=3, mode=2, stdev=2.15).
The distribution of threads with respect to the number of posts by the target user is shown in \cref{counts_threads}.
We exclude threads containing less than two posts from target users, as we cannot assess the impact of interaction between target and participating users.
Similarly, we exclude threads containing no posts from participating users.
On average, each thread contains 6.62 replies (median=6, mode=5, stdev=5.67) from 4.76 participating users (median=4, mode=4, stdev=2.45).

In \cref{distr_threads}, we report the distribution of labels for the final posts of target users in relation to the number of times target users post in a thread.
Interestingly, we observe that the more a target user engages with participating users, the less likely their final post is labeled \textit{green} (Pearson's correlation, $\rho=-0.91$, $p<0.0001$.)
After manually reviewing fifty of the longest threads in the dataset, we hypothesize that, in these cases, participating users struggle to connect with target users for a meaningful two-way conversation, thus failing to ameliorate any distress.
This suggests that in order to effectively classify the final status of target users, language and topics from target and participating users should be modeled separately, as the mental state of a target user is not only influenced by the replies they receive, but it is also expressed through the the target user's intermediate posts.

\subsection{Ethical Considerations} 
\label{sub:ethical_considerations}

Ethical considerations for our effort closely mirror those described by \citet{milne} in constructing and annotating the original dataset.
Particular care should be placed on minimizing any harm that could arise from a system deployed to notify clinicians of crises.
When an individual is identified as having a moment of crisis, direct contact might aggravate their mental state.
False alarms are also usually undesirable and can be distressful especially for people with a history of mental health struggles.
To minimize risk, additional precautions should be taken.
Examples of such measures include notifying the forum users of the automated system and explaining its purpose and functionality, and asking the users (and their clinicians) for permission to make contact during a crisis.

%% file: sections/methods.tex
\documentclass[../naaclhlt2018.tex]{subfiles}

\section{Classifying Flagged Threads}
\label{methods}

In this section, we describe the system and features we propose for classifying the final status of initially \flagged threads.
Based on the analysis of conversation dynamics in threads discussed in the previous section, we model the target and participating users in a thread separately.
As such, for every thread $t_i$ from target user $u_T$, we first partition $t_i$ in subsets $\textit{Posts}(t_i, u_T)$ of posts authored by $u_T$ and $\textit{Posts}(t_i, \overline{u_T})$ of all posts in $t_i$ authored by participating users.
Then, we extract the following identical groups of features from subsets $\textit{Posts}(t_i, u_T)$ and $\textit{Posts}(t_i, \overline{u_T})$:

\begin{itemize}[wide, labelindent=0pt]
    \item \textbf{LIWC}: We consider 93 indicators in the Linguistic Inquiry and Word Count (LIWC) dictionary \cite{pennebaker2015development}. Previous research found these features to be effective in capturing language patterns for distressed mental state \cite{kumar2015detecting,milne,cohan2017triaging,yates2017}. In contrast with other efforts, we consider LIWC features for participating users.
\item \textbf{Sentiment and subjectivity}: We consider sentiment and subjectivity of posts from target and participating users. We extract these features using the TextBlob tool\footnote{\url{https://textblob.readthedocs.io/}}.
\item \textbf{Topic modeling}: To investigate the conversation topics, we perform LDA topic modeling \cite{blei2003latent} on a subset of the CLPsych ReachOut.com dataset.
LDA analysis has been successfully used to study how users talk about their mental health conditions online \cite{resnik2015beyond}.
In particular, we use posts from the ``Wellbeing'' and ``Tough Times'' sub-forums to build a model that captures mental health topics. We exclude the ``Hang out'' and ``Introduction'' sub-forums to prevent topic drift. We used LDAvis \cite{sievert2014ldavis} to inspect and tune the number of topics. We tested models with 5 to 50 topics, and ultimately settled on 10 through empirical evaluation.
We use the LDA implementation of Gensim\footnote{\url{https://radimrehurek.com/gensim/}} to compute the model.
\item \textbf{Textual features}: To more precisely capture the content of posts, we consider single-word tokens extracted from posts as features. We remove stopwords, numbers, and terms appearing in less than 5 posts. When representing posts, we weight terms using \textit{tf-idf}.
\end{itemize}

Besides the sets of features shared between target and participating users, we also consider the following signals as features:
likelihood of the first reply in the thread being \green, as assigned by the classifier by \citet{cohan2017triaging};
number of posts in the thread from the target user;
number of posts in the thread from participating users;
number of posts in the thread from a moderator.

We experiment with several classification algorithms, which we compare in \cref{classification}, while we present the outcome of our feature analysis in \cref{feature_analysis}.

%% file: sections/results.tex
\documentclass[../naaclhlt2018.tex]{subfiles}

\section{Results}
\label{sec:results}

\subsection{Classification}
\label{classification}

We report results of the final state classification in \cref{prf1}. We train all methods shown here on all features described in \cref{methods} and we test with five-fold cross validation.
We observe that all methods perform better than a majority classifier baseline.
In particular, XGBoost \cite{chen2016xgboost} achieves the best precision, but a simple logistic regression model with ridge penalty achieves better F1 score. This suggests that the former might suffer from overfitting due to the limited size of the training data.
The classification outcome of the logistic regression model is statistically different than all other models (Student $t$-test, $p < 0.001$, Bonferroni-adjusted.)

\begin{table}[t]
\centering
\small
\renewcommand*{\arraystretch}{1.3}

\begin{tabular}{@{}lrrr@{}}
\toprule
\textbf{Method}                    & \textbf{Pr}    & \textbf{Re}    & \textbf{F1}    \\ \midrule
Majority classifier & 37.69          & 50.00          & 42.98          \\
Logistic regression & 71.64          & \textbf{68.16} & \textbf{69.10} \\
Linear SVM          & 70.40          & 61.41          & 62.83          \\
XGBoost             & \textbf{75.72} & 63.95          & 66.06          \\ \bottomrule
\end{tabular}
\caption{Performance of several classification methods of determining the final state of a target user in a \flagged thread.
The logistic regression is statistically different from all other models (Student $t$-test, $p < 0.001$, Bonferroni-adjusted.)}
\label{prf1}
\end{table}

\begin{figure}[t]
    \centering
    \includegraphics[width=.98\columnwidth]{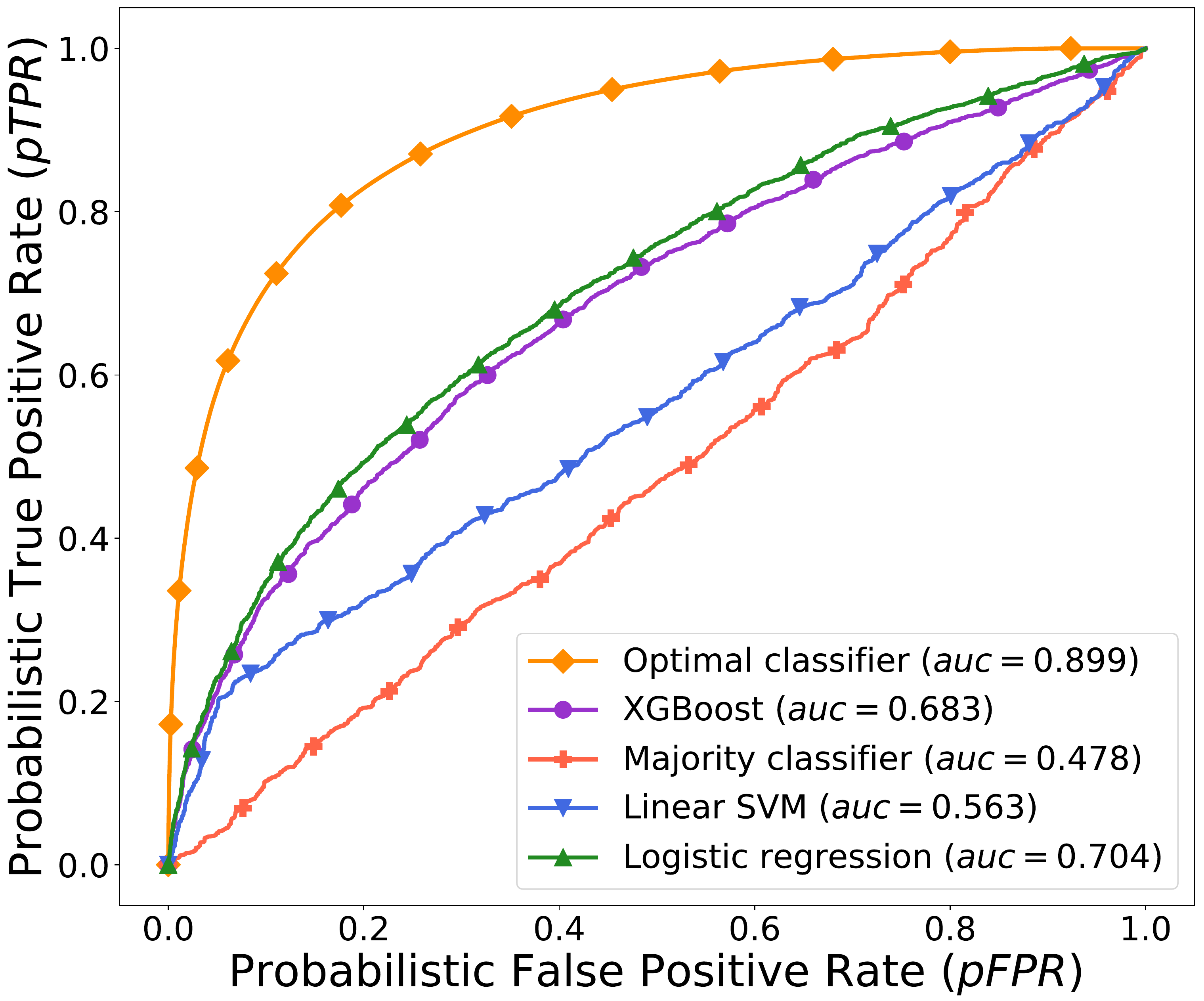}
    \caption{Probabilistic receiving operating characteristic \cite{burl1994automated} for the classification methods.
    Because the labels obtained from \cite{cohan2017triaging} are real number in the range $\lbrack0, 1\rbrack$, results evaluated on them represent lower bounds on performance of classifiers.
    The optimal ROC achievable by any classifier is shown in orange (optimal $AUC=0.899$.)}
    \label{proc}
\end{figure}

In \cref{proc}, we report results of a variant of receiving operating characteristic (ROC) analysis designed to handle probabilistic labels.
Recall that the labels obtained from \citet{cohan2017triaging} are real numbers in the range $\lbrack0, 1\rbrack$ representing the likelihood of each post of being \textit{green}.
While the labels can be turned into binary labels (as done to compute precision, recall, and F1 score reported in \cref{prf1}), doing so ignores the uncertainty associated with probabilistic labels.
Techniques to modify ROC analysis to consider probabilistic labels have been proposed in the literature. We consider the variant introduced by \citet{burl1994automated} which was recently used to evaluate cohort identification in the medical domain from web search logs \citep{soldaini2017inferring}.
Results in \cref{proc} largely mimic performance shown in \cref{prf1}, with the logistic regression model outperforming all other classifiers and achieving $78.3\%$ of the area under the curve (AUC) of the optimal classifier (i.e., a classifier that always predicts the exact value of probabilistic labels.)
While the ROC curve for the XGBoost is comparable to the logistic regression classifier, we observe that the SVM model achieves similar performance to the two only for high confidence samples (bottom left corner), and its performance declines sharply when more positive samples are inferred.

\begin{table}[t]
\centering
\small
{
\renewcommand*{\arraystretch}{1.0}
\begin{tabular}{@{}lrrr@{}}
\toprule
\textbf{Feature set} & \textbf{Pr} & \textbf{Re} & \textbf{F1} \\ \midrule
\begin{tabular}[c]{@{}l@{}}Only features from \\ target users\end{tabular}                                   & 69.72       & 62.40       & 63.93       \\[7mm]
\begin{tabular}[c]{@{}l@{}}Averaged features \\ from target and \\ participating  users\end{tabular}          & 60.04       & 58.90       & 59.31       \\[8mm]
\begin{tabular}[c]{@{}l@{}}Separate symmetric \\ features for target and \\ participating users\end{tabular} & \textbf{71.64}       & \textbf{68.16}       & \textbf{69.10}       \\ \bottomrule
\end{tabular}}
\caption{Comparison of different strategies for extracting features from target and participating users. The best feature set (separate symmetric features) is significantly better than the other two (Student \textit{t}-test, $p < 0.001$, Bonferroni-adjusted.)}
\label{combination}
\end{table}

Finally, results shown in \cref{combination} motivate our approach for separately modeling features for target and participating users.
We observe that using separate symmetric features for the two groups of users not only improves upon using features from target user posts alone, it also outperforms averaging features extracted from the two groups together (Student t-test, $p < 0.001$, Bonferroni-adjusted.)
This empirically confirms our hypothesis that language and topics from target and participating users should be modeled separately.

\subsection{Features analysis}
\label{feature_analysis}

We report the result of an ablation study on feature groups in \cref{ablation}.
For all runs, we always include the group of ``shared'' features detailed in \cref{methods}.
Overall, we observe that the method of including all feature sets outperforms all other runs. We note, however, that the addition of most feature sets only sums to a modest improvement (up to 7.6\% in F1 score) over LIWC features alone, which confirms previous observations on the effectiveness of LIWC in modeling mental health language \cite{kumar2015detecting,milne,cohan2017triaging,yates2017}.

Beside LIWC, we found LDA and sentiment features to be moderately effective for user mental state classification.
On the other hand, we found token features to have a limited impact on the performance of the system, improving the overall F1 score by just 0.35\%.
Their contribution was also found to be insignificant (Student \textit{t}-test, $p=0.19$.)
We attribute this result to the fact that, compared to other features, non-sentiment terms used by target or participating users represent a much weaker signal for modeling the change in self-harm risk that interests us.

\begin{table}[t]
\centering
\small
{
\renewcommand*{\arraystretch}{1.0}
\begin{tabular}{@{}lrrr@{}}
\toprule
\textbf{Feature group}                                                       & \textbf{Pr}    & \textbf{Re}    & \textbf{F1}    \\ \midrule
LIWC                                                                         & 67.85*          & 63.12*          & 64.26*          \\[4mm]
\begin{tabular}[c]{@{}l@{}}LIWC + Sentiment \end{tabular}                                                             & 68.34*          & 63.80*          & 64.93*          \\[4mm]
\begin{tabular}[c]{@{}l@{}}LIWC + Sentiment  + \\ LDA topics \end{tabular}            & 70.95           & 67.73 & 68.83 \\[5mm]
\begin{tabular}[c]{@{}l@{}}LIWC + Sentiment  + \\ LDA topics + Tokens\end{tabular} & \textbf{71.64} & \textbf{68.16} & \textbf{69.10} \\ \bottomrule
\end{tabular}
}
\caption{Ablation study of feature groups.
Results marked by * indicate runs that are significantly different from the best method (Student \textit{t}-test, $p < 0.001$, Bonferroni-adjusted.)}
\label{ablation}
\end{table}

We report the most significant features for each feature group in \cref{bestworst-features}.
For each group of features, we report the top three positive and negative features for target and participating users.
In order to improve readability, feature weights are $\ell_2$-normalized with respect to their group (token features are one to two orders of magnitude smaller than the other groups.)
For LDA features, we report a list of significant terms for topics, as well as possible interpretations of them, in \cref{lda}.

When analyzing LIWC features for the target users, we note that mention of support communities (e.g., religion), internet slang (netspeak), and talk about leisure activities correlate with a decrease in risk by the end of a thread.
Conversely, filler language (which can sometimes indicate emotional distress \cite{stasak77depression}), mention of family, and swearing are all associated with target users remaining in a \flagged status at the end of a thread.
While participants share some positive LIWC features with target users (e.g., religion), we notice that mention of home and family are, in this case, associated with a positive outcome.
To explain this difference, we sampled 20 threads in which target or participating users mentioned ``family'' or ``home.'' We empirically observe that, in a majority of cases, when target users mention family it is because they have trouble communicating with or relating to them.
On the other hand, when participating users mention family and home it is usually to remind target users of their relationships with loved ones.
While not conclusive, this observation suggests a possible explanation of the difference.

\begin{table*}[th]
\centering
{\scriptsize
\renewcommand*{\arraystretch}{1.2}

\begin{tabular}{rlrlrlrlrlrl}
\toprule
\multicolumn{4}{c}{\textbf{LIWC}} & \multicolumn{4}{c}{\textbf{LDA}} & \multicolumn{4}{c}{\textbf{Tokens}} \\
\multicolumn{2}{c}{\textbf{target users}} & \multicolumn{2}{c}{\textbf{participants}} & \multicolumn{2}{c}{\textbf{target user}} & \multicolumn{2}{c}{\textbf{participants}} & \multicolumn{2}{c}{\textbf{target users}} & \multicolumn{2}{c}{\textbf{participants}} \\ \midrule
+0.423 & religion & +0.528 & home & +0.471 & topic \#4 & +0.397 & topic \#8 & +0.847 & \textit{thanks} & +0.541 & \textit{proud} \\
+0.339 & netspeak & +0.478 & family & +0.245 & topic \#8 & +0.303 & topic \#6 & +0.614 & \textit{hope} & +0.521 & \textit{value} \\
+0.295 & leisure & +0.436 & religion & +0.210 & topic \#7 & +0.272 & topic \#3 & +0.570 & \textit{didn't} & +0.509 & \textit{dreams} \\
\multicolumn{2}{c}{...} & \multicolumn{2}{c}{...} & \multicolumn{2}{c}{...} & \multicolumn{2}{c}{...} & \multicolumn{2}{c}{...} & \multicolumn{2}{c}{...} \\
-0.126 & filler & -0.447 & swear & -0.624 & topic \#9 & -0.756 & topic \#10 & -0.994 & \textit{anymore} & -0.582 & \textit{ready} \\
-0.392 & family & -0.333 & sexual & -0.217 & topic \#3 & -0.274 & topic \#7 & -0.651 & \textit{scared} & -0.572 & \textit{trying} \\
-0.354 & swear & -0.248 & money & -0.092 & topic \#1 & -0.088 & topic \#1 & -0.598 & \textit{I'm} & -0.477 & \textit{school} \\ \bottomrule
\end{tabular}}
\caption{Top positive and negative features for each feature group. Scores are $\ell_2$-normalized with respect to their group. For LDA, we report a list of significant terms for each topic, as well as possible interpretations, in \cref{lda}.}
\label{bestworst-features}
\end{table*}

\begin{table}[th]
\centering
{\scriptsize
\renewcommand*{\arraystretch}{1.7}
\begin{tabular}{cll}
\toprule
\textbf{Topic} & {\renewcommand*{\arraystretch}{0.9}\textbf{\begin{tabular}[c]{@{}l@{}}Significant terms by relevance\\ \cite{sievert2014ldavis}, $\lambda = 0.6$\end{tabular}}} & {\renewcommand*{\arraystretch}{0.9}\textbf{\begin{tabular}[c]{@{}l@{}}Potential\\ interpretation\end{tabular}}} \\ \midrule
1 & \textit{like, think, life, need, know, will, i've, feeling} & ? \\
2 & \textit{help, people, reachout, talk, need, support} & {\renewcommand*{\arraystretch}{0.9}\begin{tabular}[c]{@{}l@{}} discussions \\about ReachOut  \end{tabular}} \\
3 & \textit{important, home, people, family, need, back} & family \\
4 & \textit{think, time, want, now, today, people, help} & ? \\
5 & \textit{know, feel, great, negative, day, work, person} & ? \\
6 & \textit{didn't, who, like, know, feeling, will, life} & ? \\
7 & \textit{thanks, want, life, feel, hey, great, guys} & {\renewcommand*{\arraystretch}{0.9}\begin{tabular}[c]{@{}l@{}}thanking other \\ReachOut users \\ for their support \end{tabular}} \\
8 & \textit{talk, school, time, self, counsellor, seeing} & therapy, school \\
9 & \textit{work, time, paycheck, study, uni, goal} & work, study \\
10& \textit{see, body, hard, care, health, weight} & weight loss \\ \bottomrule
\end{tabular}}
\caption{
    Significant terms for the LDA topics computed over the ReachOut forum.
    Significance is determined using \textit{relevance} \citep{sievert2014ldavis}.
}
\label{lda}
\end{table}

Compared to LIWC categories, we found the scores assigned to LDA topics more difficult to explain.
As shown in \cref{lda}, while some topics have clearly defined subjects, others are harder to interpret.
However, we note that most topics reported in \cref{bestworst-features} as having a high positive or negative weight for our best classifier have a clear interpretation.
For example, topics \#8, \#7, and \#3 are about school and counseling, thanking the ReachOut.com community, and family.
Among negative topics, discussion of weight loss (topic \#10), or work and university (topic \#9) are associated with \flagged states.
Interestingly, topic \#7 has a positive weight for target users (this is expected: users who thank other users for their help are more likely to have transitioned to \green by the end of a thread), but it has a negative weight for participants.
We could not find a plausible intuition for the latter observation.
Similar to LIWC features, we found that family is correlated with a decrease in risk by the end of a thread when mentioned by participating users (topic \# 3), while the opposite is true for target users.

We analyze the importance values associated with tokens.
We note that observations of these features are less likely to be conclusive, given their limited impact on classification performance (\cref{ablation}.)
Nevertheless, we observe that features in this category either represent emotional states that are also captured by sentiment features (e.g., \textit{hope, proud, scared}) or relate to topics discussed in the previous section (e.g., \textit{school}, \textit{thanks}.)

Finally, while not reported in \cref{bestworst-features}, we also study the importance of sentiment and subjectivity features.
We observe that sentiment positively correlates with both target and participating users ($+0.902$ and $+0.629$, respectively).
High levels of objectivity by target users correlate with a decrease in risk of harm by the end of a thread ($+0.510$), while the model finds that objectivity by participating users is not predictive of a \green final state (assigned weight: $+0.033$.)

\subsection{Conversation Threads Analysis} 
\label{sub:qualitative_analysis}

Beside analyzing individual features, we also present an overview of how aspects of threads correlate with performance of the classifier.
We observe that there exists a positive correlation between the length of a thread and the performance of the classifier.
Such correlation is significant for both \green ($\rho=0.33$, $p < 0.05$) and \flagged ($\rho=0.37$, $p < 0.05$) conversation threads, and likely explained by observing that longer threads may contain more information about the mental state of target users.
We note that the average standard deviation of target user state within a thread is $0.32$ (median=$0.19$), which suggests that some target users oscillate between \green and \flagged states in a conversation.
However, we observe no correlation between variance and model performance (\textit{p}-value$=0.44$.)

While encouraging, we recognize the limitations of our current approach.
Online data is a great resource to study the language of mental health, but it often lacks granularity.
This is not an issue for long-trend studies, but it poses issues when trying to model language around suicidal ideation, given the short duration of suicidal crises \cite{hawton2007restricting}.
While analyzing conversations that were incorrectly classified by our system, we also noted that several target users transitioned between states without any meaningful interaction at all with participating users.
Our intuition is that the mental state of a target user is significantly influenced by passively reading other threads, interacting over a secondary channel like private messages, and experiencing offline events, none of which are available as inputs to our system, thereby limiting its accuracy.

%% file: sections/conclusions.tex

\documentclass[../naaclhlt2018.tex]{subfiles}

\section{Conclusions}
\label{sec:conclusions}

In recent years, the number of online communities designed to offer support for mental health crises has grown significantly.
While users generally find these communities helpful, researchers have shown that, in some cases, they could also normalize harmful behavior, discourage professional treatment, and instigate suicidal ideation.
We study the problem of assessing the impact of interaction with a support community on users who are suicidal or at risk of self-harm.

First, using the 2016 CLPsych ReachOut.com corpus, we build a dataset of conversation threads between users in distress and other members of the community.
Then, we construct a classifier to predict whether an at-risk user can successfully overcome their state of distress through conversations with other community members.
The classifier leverages LIWC, sentiment, topic, and textual features.
On the dataset introduced in this paper, it achieves a $0.69$ macro-F1 score.
Furthermore, we analyze the effectiveness of features from our model to gain insights from the language used and the topics appearing in conversations with at-risk users.